\documentclass{article} % For LaTeX2e
\usepackage[final]{colm2025_conference}

\usepackage{microtype}
\usepackage{hyperref}
\usepackage{url}
\usepackage{booktabs}
\usepackage[utf8]{inputenc} % allow utf-8 input
\usepackage[T1]{fontenc}    % use 8-bit T1 fonts
\usepackage{hyperref}       % hyperlinks
\usepackage{url}            % simple URL typesetting
\usepackage{booktabs}       % professional-quality tables
\usepackage{amsfonts}       % blackboard math symbols
\usepackage{nicefrac}       % compact symbols for 1/2, etc.
\usepackage{microtype}      % microtypography
\usepackage{xcolor}         % colors
\usepackage{colortbl}
\usepackage{graphicx}
\usepackage{amsmath}
\usepackage{makecell}
\usepackage{multirow}
\usepackage{pifont}
\usepackage[toc,page,header]{appendix}
\usepackage{minitoc}
\usepackage{wrapfig}
\usepackage{csquotes}
\usepackage{subcaption}
\usepackage{caption}
\usepackage{adjustbox}
\usepackage{colortbl}
\usepackage[most]{tcolorbox}
\usepackage{lineno}
\tcbset{colback=white}
\DeclareCaptionFormat{nosublabel}{#3}
\DeclareCaptionLabelFormat{nosublabel}{}
\usepackage{pgf}
\usepackage{pgfplots}

\newcommand{\cmark}{\ding{51}}
\newcommand{\xmark}{\ding{55}}

\newcommand{\applyGradientRaven}[1]{%
    \ifdim #1 pt > 80pt \cellcolor{black!3}{#1}%
    \else\ifdim #1 pt > 24pt \cellcolor{black!7}{#1}%
    \else\ifdim #1 pt > 18pt \cellcolor{black!12}{#1}%
    \else\ifdim #1 pt > 12.5pt \cellcolor{black!18}{#1}%
    \else\cellcolor{black!25}{#1}%
    \fi\fi\fi\fi%
}

\newcommand{\applyGradientOther}[1]{%
    \ifdim #1 pt > 60pt \cellcolor{black!3}{#1}%
    \else\ifdim #1 pt > 40pt \cellcolor{black!7}{#1}%
    \else\ifdim #1 pt > 30pt \cellcolor{black!12}{#1}%
    \else\ifdim #1 pt > 25pt \cellcolor{black!18}{#1}%
    \else\cellcolor{black!25}{#1}%
    \fi\fi\fi\fi%
}

\tcbset{
  aibox/.style={
    width=\linewidth,
    top=10pt,
    colback=white,
    colframe=black,
    colbacktitle=black,
    enhanced,
    center,
    attach boxed title to top left={yshift=-0.1in,xshift=0.15in},
    boxed title style={boxrule=0pt,colframe=white,},
  }
}
\newtcolorbox{AIbox}[2][]{aibox,title=#2,#1}

\definecolor{darkgreen}{RGB}{0,130,0}
\definecolor{darkred}{RGB}{180,0,0}

\title{What is the Visual Cognition Gap between Humans and Multimodal LLMs?}

\author{Xu Cao$^{1*}$,  Yifan Shen$^{1*}$,  Bolin Lai$^{2}$,  Wenqian Ye$^{3}$,  Yunsheng Ma$^{4}$,  Joerg Heintz$^{1}$ \\ \textbf{Jintai Chen}$^{5}$,  \textbf{Meihuan Huang}$^{6}$,  \textbf{Jianguo Cao}$^{6}$,  \textbf{
Aidong Zhang}$^{3}$,  \textbf{James M. Rehg}$^{1}$ \\
    $^{1}$Department of Computer Science, University of Illinois at Urbana-Champaign \\
    $^{2}$College of Computing, Georgia Institute of Technology \\
  $^{3}$Department of Computer Science, University of Virginia \\
  $^{4}$Digital Twin Lab, Purdue University \\
  $^{5}$HKUST (Guangzhou) \\
  $^{6}$Department of Rehabilitation Medicine, Shenzhen Children's Hospital \\
  \texttt{\{xucao2,yifan26,jrehg\}@illinois.edu}
}

\begin{document}

\maketitle
\let\thefootnote\relax\footnotetext{$^*$Equal contribution.}
\begin{abstract}
Recently, Multimodal Large Language Models (MLLMs) and Vision Language Models (VLMs) have shown great promise in language-guided perceptual tasks such as recognition, segmentation, and object detection. However, their effectiveness in addressing visual cognition problems that require high-level multi-image reasoning and visual working memory is not well-established. One such challenge is matrix reasoning -- the cognitive ability to discern relationships among patterns in a set of images and extrapolate to predict subsequent patterns. This skill is crucial during the early neurodevelopmental stages of children. Inspired by the matrix reasoning tasks in Raven’s Progressive Matrices (RPM) and Wechsler Intelligence Scale for Children (WISC), we propose a new dataset MaRs-VQA to evaluate the visual cognition capability of MLLMs and compare their performance with existing human visual cognition studies. Based on the training data of MaRs-VQA, we also finetune a baseline model Qwen2-VCog with multi-stage cognition reasoning annotations. Our comparative experiments with different baselines reveal a gap between MLLMs and human intelligence, highlighting the visual cognitive limitations of current MLLMs. We believe that the public release of MaRs-VQA and the Qwen2-VCog baseline model will drive progress toward the next generation of MLLMs with human-like visual cognition abilities. MaRs-VQA is available at \href{https://huggingface.co/datasets/IrohXu/VCog-Bench}{huggingface.co/datasets/IrohXu/VCog-Bench}. The training code of Qwen2-VCog is available at \href{https://github.com/IrohXu/Cognition-MLLM}{github.com/IrohXu/Cognition-MLLM}.
\end{abstract}

\section{Introduction}
\label{sec:introduction}

\begin{figure}[!t]
    \vspace{-0.5cm}
    \centering
    \includegraphics[width=1.0\linewidth]{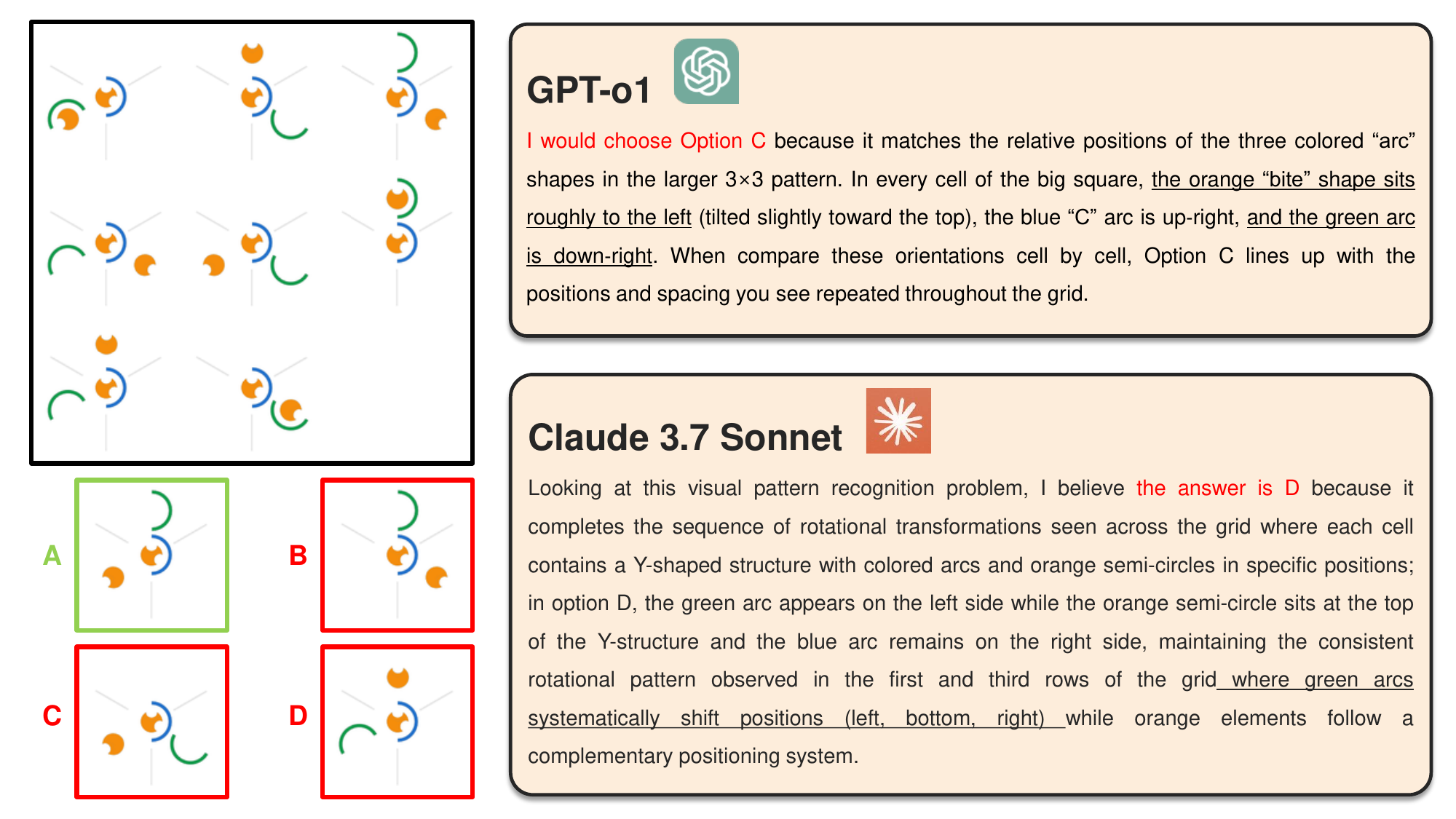}
    \caption{The example of the subpar performance of current state-of-the-art MLLMs (GPT-o1, Claude 3.7 Sonnet) on a simple matrix reasoning task used in MaRs-VQA (similar to cases in RPM and WISC). Both models can recognize the basic shapes in the provided patterns but fail to reason the next pattern.}
    \label{fig:example_gpt4v}
    \vspace{-0.5cm}
\end{figure}

\textbf{Matrix reasoning} is a crucial ability in human cognition. It is used in non-verbal, culture-reduced intelligence assessments because it minimizes the influence of acquired knowledge and skills~\citep{jensen1998factor,jaeggi2010relationship,laurence2023cognitive}. Common matrix reasoning problems consist of images with simple symbols governed by underlying abstract rules~\citep{malkinski2023review} (see Figure~\ref{fig:example_gpt4v}). Participants have to identify and comprehend the rules based on a few provided patterns, and then reason about the next pattern following the same rules. Matrix reasoning is an important reflection of many fundamental capabilities of human visual cognition, such as processing speed and working memory, that emerge in the early stages of children’s neurodevelopment~\citep{gentner1977children}. It is also included in many assessment methods for fluid intelligence tests such as Wechsler Intelligence Scale for Children (WISC)~\citep{wechsler1949wechsler} and Raven's Progressive Matrices (RPM)~\citep{raven2003raven}.

In psychometrics, matrix reasoning tasks for children are specifically designed to assess visual reasoning abilities without prior specialized training. Children typically approach these tests relying solely on their general cognitive skills developed from everyday interactions with natural environments. This raises an intriguing question: Do Multimodal Large Language Models (MLLMs) exhibit visual cognitive capabilities similar to those of humans? MLLMs are trained on extensive general-domain data and have demonstrated the ability to generalize to unfamiliar tasks through in-context learning. However, current MLLMs still struggle with tasks that require advanced inductive reasoning, as evidenced by their poor performance on abstract reasoning tests such as the RAVEN IQ-test~\citep{huang2024language,fu2024blink,yiu2024kiva}. The RAVEN IQ-test itself has notable limitations, including a relatively small dataset of only 50 samples~\citep{huang2024language}, potentially introducing randomness and failing to robustly evaluate MLLMs' capabilities. Furthermore, it lacks comparative analyses with human performance, underscoring the need for more comprehensive and rigorous evaluation methods.

To address these gaps, we propose MaRs-VQA, a new visual question answering (VQA) dataset for general-purpose MLLMs, based on psychologist-certified matrix reasoning items with extensive sample diversity and rigorous human reference~\cite{chierchia2019matrix}. Unlike recent works, our benchmark uniquely offers: (i) direct comparison between generic MLLMs and humans on a scale an order of magnitude larger than previous VQA benchmarks and with richer stimuli types; (ii) rigorous psychological validity and baseline from human subject studies; (iii) explicit dual-modality annotation---for every question, both image and natural-language descriptions are provided for all options, enabling probing of language-vs-visual inference; (iv) full chain-of-thought (CoT) reasoning steps annotated, supporting deeper cognitive diagnosis and fine-tuning. 
% These features allow MaRs-VQA to serve as a diagnostic tool for visual cognitive capacity, setting it apart from both rule-based and black-box "IQ" benchmarks. This is the largest psychologist-verified VQA dataset for matrix reasoning assessment including 1,440 examples in total. The sample diversity of MaRs-VQA also surpasses other datasets before. It contains over 50 types of shape, 16 types of colour and over 500 graphic combinations. 
 
We also conduct thorough evaluation and comparison across 5 existing MLLMs (including their variants) and human performance under zero-shot inference setting (no prior knowledge) on MaRs-VQA and another abstract reasoning datasets RAVEN containing human studies. In our experiments, we observe that MLLMs with more parameters generally perform better on our benchmark, adhering to established scaling laws in a limited scope. However, even the largest open-source MLLMs and GPT-4o fall short of surpassing human performance in matrix reasoning tasks. In conclusion, our contributions are summarized as follows:

\begin{itemize}
    \item We introduce a new matrix reasoning VQA dataset -- MaRs-VQA, containing 1,440 image instances designed by psychologists, which is the largest dataset for matrix reasoning zero-shot evaluation.
    \item We conducted supervised fine-tuning (SFT) of Qwen2-VL using our annotated cognitive reasoning data. Our results indicate that fine-tuning enhances Qwen2-VL's accuracy on MaRs-VQA to match human performance levels; however, its generalization capabilities remain limited.
    \item Our thorough experiments qualitatively reveal the visual cognition gap between MLLMs and humans in matrix reasoning problems. We also show additional insights of deficiencies in MLLMs, which can inspire more future investigations in model design.
\end{itemize}

\section{Related Works}
\label{sec:related_works}

\paragraph{Cognitive Test of Large Language Models (LLMs)}

The rise of LLMs has aroused interest in exploring human-like AI in psychology and cognition~\citep{ullman2023large}. Recent works tested LLMs' cognitive abilities in causal reasoning~\citep{binz2023using}, abstract reasoning~\citep{xu2023llms,moskvichev2023conceptarc,jiang2024marvel,ahrabian2024curious}, analogical reasoning~\citep{webb2023emergent},  systematic reasoning~\citep{hagendorff2023human}, and theory of mind~\citep{strachan2024testing}. Despite this success, only limited research has been conducted on the areas of MLLMs and visual cognition. Visual cognition involves the process by which the human visual system interprets and makes inferences about a visual scene using partial information. It is observed that while LLMs demonstrate a basic understanding of physical laws and causal relationships, they lack deeper insights into intuitive human preferences and reasoning~\cite{buschoff2023have}. Almost all existing visual cognition benchmarks focus on testing MLLMs' cognitive abilities in simple tasks~\citep{zhou2023mental,jassim2023grasp}, and ignore testing complex abstract reasoning and logical reasoning ability. Therefore, new and challenging benchmarks based on the theory of visual cognition are needed to assess and improve AI systems' capabilities for human-like visual understanding.

\paragraph{Matrix Reasoning}

Matrix reasoning is often used to determine human intelligence related to visual cognition and working memory~\citep{salthouse1993influence,jaeggi2010relationship,fleuret2011comparing} that is widely used by RPM~\citep{raven2003raven,soulieres2009enhanced}, WISC~\citep{wechsler1949wechsler,kaufman2015intelligent} to evaluate human ability to detect the underlying conceptual relationship among visual objects and use reasoning to find visual cues. Early research indicated that deep learning models can be trained to solve simple matrix reasoning~\citep{malkinski2022deep,malkinski2023review,xu2023abstract,malkinski2024one} and compositional visual relation tasks~\citep{fleuret2011comparing,zerroug2022benchmark,liu2021learning}. Several datasets and benchmarks are also proposed, such as RAVEN~\citep{zhang2019raven}, RAVEN-I~\citep{hu2021stratified}, RAVEN-FAIR~\citep{benny2021scale}, CVR~\citep{zerroug2022benchmark}. However, these works have a key limitation. They overlook the fact that humans can solve such puzzles in a zero-shot manner, without explicit training on large-scale data. Recently, there are also some useful zero-shot visual reasoning inference datasets such as RAVEN-IQ~\citep{huang2024language}, Visual Reasoning Benchmark~\citep{zhang2024far}, but all of them are limited by lacking rigorous human experiments as reference and conducting experiments on relatively small datasets without psychometrical validation.

\paragraph{Vision-Language Models}

Researchers have been actively investigating the utility of Vision-Language Models (VLMs) for addressing vision reasoning tasks~\citep{zellers2019recognition,bordes2024introduction}. These latest VLMs are constructed using a combination of the CLIP vision encoder, pretrained LLMs, and a connected adapter to align visual features with language space~\citep{zhang2024mm,shao2024visual,gupta2023visual,fu2024blink}. Notably, methodologies such as MiniGPT-4~\citep{zhu2023minigpt}, InstructBLIP~\citep{dai2024instructblip}, LLaVA~\citep{liu2024visual}, CogVLM~\citep{wang2023cogvlm} underscore the significance of employing high-quality visual instruction tuning data. Nevertheless, current VLMs encounter challenges in adapting to high-resolution and visually complex images. These problems stem from the absence of a robust visual search mechanism~\citep{wu2023textit}, few-shot reasoning~\citep{guo2023large}, compositional understanding~\citep{yuksekgonul2022and} and the constrained visual grounding capabilities inherent in CLIP~\citep{tong2024eyes}.

\section{MaRs-VQA Dataset}

\begin{table}[!t]\footnotesize
  \vspace{-0.4cm}
  \centering
  \adjustbox{max width=\textwidth}{
  \begin{tabular}{ccccccccc}
    \toprule
    Dataset & Source & Sample & Instance & \thead{\tiny RGB image}  & \thead{\tiny Human \\ \tiny Study} &  \thead{\tiny Psychological \\ \tiny Validity} & \thead{\tiny Open-source} & \thead{\tiny VQA \\ \tiny Annotation}\\
    \midrule \midrule
    \thead{\scriptsize kosmos-iq50 (NeurIPS-23) \\ \scriptsize \citep{huang2024language}}
    & \thead{\scriptsize RAVEN-IQ Test} &
    \begin{minipage}[b]{0.2\columnwidth}
        \centering
        \raisebox{-.45\height}{\includegraphics[width=0.6\linewidth]{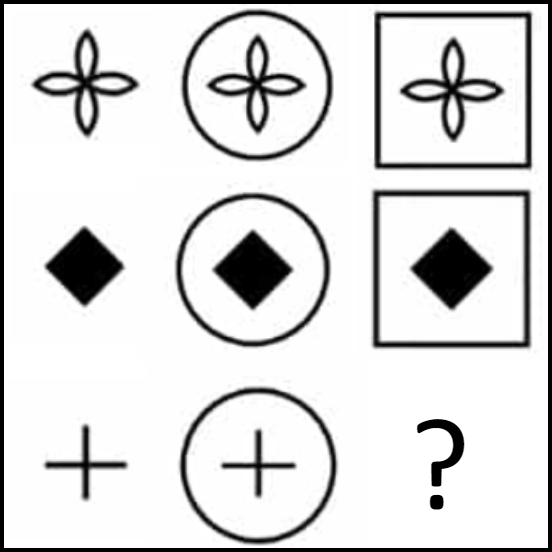}}
    \end{minipage}
    & 50
    & \color{darkred}\xmark\color{black} & \color{darkred}\xmark\color{black} & \color{darkgreen}\cmark\color{black} & \color{darkred}\xmark\color{black} & \color{darkred}\xmark\color{black}  \\
    \midrule
    \thead{\scriptsize Visual Reasoning Benchmark \\ \scriptsize (COLM-24)~\citep{zhang2024far}}
    & \thead{\scriptsize Mensa Test, RAVEN, \\ \scriptsize IntelligenceTest} &
    \begin{minipage}[b]{0.2\columnwidth}
        \centering
        \raisebox{-.45\height}{\includegraphics[width=0.6\linewidth]{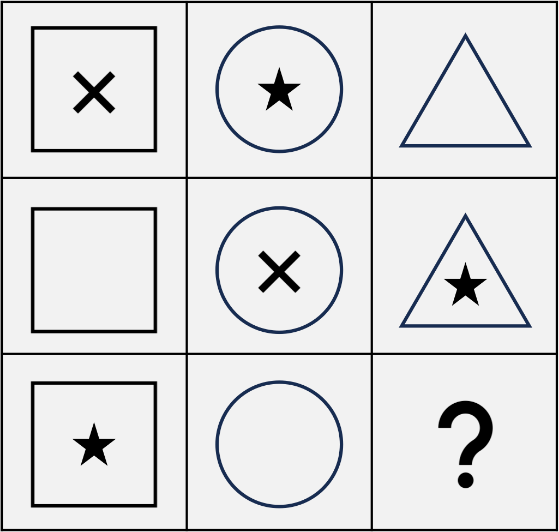}}
    \end{minipage}
    & 241
    & \color{darkred}\xmark\color{black} & \color{darkred}\xmark\color{black} & \color{darkred}\xmark\color{black} & \color{darkred}\xmark\color{black} & \color{darkred}\xmark\color{black}  \\
    \midrule
    \thead{\scriptsize MaRs-VQA \\ \scriptsize (ours)}
    & {\scriptsize MaRs-IB Questionnaire} &
    \begin{minipage}[b]{0.2\columnwidth}
        \centering
        \raisebox{-.45\height}{\includegraphics[width=0.6\linewidth]{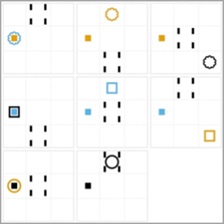}}
    \end{minipage}
    & 1,440
    & \color{darkgreen}\cmark\color{black} & \color{darkgreen}\cmark\color{black} & \color{darkgreen}\cmark\color{black} & \color{darkgreen}\cmark\color{black} & \color{darkgreen}\cmark\color{black} \\
    \bottomrule
  \end{tabular}
  }
  \caption{Comparison of recently released zero-shot matrix reasoning datasets to evaluate MLLMs.}
  \label{tab:dataset_comparison}
\end{table}

The MaRs-VQA dataset is designed to evaluate the zero-shot abstract reasoning capabilities of MLLMs through various matrix reasoning VQA tasks. The images in MaRs-VQA are sourced from the questionnaires in Matrix Reasoning Item Bank, which is created by psychologists including 18 sets of abstract reasoning questionnaires (80 instances in each set) for non-verbal abstract reasoning assessment of adolescents and adults~\citep{chierchia2019matrix}. Each item presents an incomplete $3 \times 3$ matrix of abstract shapes, requiring participants to identify relationships among the shapes. Then, we create VQA annotations in the images from all questionnaires. The comparison among MaRs-VQA and previous matrix reasoning benchmark datasets is shown in Table~\ref{tab:dataset_comparison}.

To transform the matrix reasoning problem into a VQA task, we define two option sets -- image-based set and language-based set. In the image-based set, we provide four candidates to the missing patch in the question. We further diversify the modalities of our dataset to support the evaluation of different kinds of models. Specifically, the author teams annotate language-based descriptions for each option, forming language-based set. Each option annotation is formatted by GPT-4o to ensure consistency. In this process, we first manually design 10 formatted language-based sample pairs. These examples are then used as few-shot samples to query GPT-4o through in-context learning. The context generation system prompt guides GPT-4o to re-caption all annotations. After generating all samples, human annotators in the author team review each option again and revise the incorrect description. In Table~\ref{tab:dataset_comparison}, compared with other matrix reasoning datasets for MLLM's visual cognition evaluation, MaRs-VQA is the largest one with unique features on psychological validity, human study reference, VQA annotations.

\section{Problem Statement}

In this section, we introduce the evaluation pipeline of MaRs-VQA under multi-image reasoning setting.

Assume that the test set contains $n$ VQA samples, denoted as $\{ (\mathbf{q}_{i}, \mathbf{x}_{i}, \mathbf{y}_i)\}^{n}_{i=1}$. $\mathbf{q}_{i}$ represents the question image showing the $3\times3$ matrix reasoning task. $\mathbf{x}_{i}=[ x^{1}_{i}, ...,  x^{k}_{i}]$ represents the images in the option set, where $k$ is the number of options. $\mathbf{y}_{i}$ is the answer of the matrix reasoning question. The inference pipeline can be formulated as:

\vspace{-0.2cm}
\begin{equation}
\begin{aligned}
    \hat{\mathbf{y}_{i}} = F_{\theta}(\mathbf{q}_{i}, \mathbf{x}_{i}, \mathbf{x}_{sys}).
\end{aligned}
\label{eq:define}
\end{equation}

$\mathbf{x}_{sys}$ is the system prompt, including independent information about the matrix reasoning problem setting, step-by-step reasoning examples (optional), few-shot examples (optional), requirements for the output format. $\hat{\mathbf{y}_{i}}$ is the prediction result. $F_{\theta}$ is an autoregressive decoder in the MLLM for answer generation. It is defined as:

\begin{equation}
\begin{aligned}
    &P(\hat{\mathbf{y}}_{i} | \mathbf{q}_{i}, \mathbf{x}_{i}, \mathbf{x}_{sys}) = &\prod_{j=1}^{L} P(\hat{\mathbf{y}}_{i,j} | f(\mathbf{q}_{i}, \mathbf{x}_{i}), \mathbf{x}_{sys}, \hat{\mathbf{y}}_{i,<j}; \theta),
\end{aligned}
\label{eq:mllm}
\end{equation}

where $f$ is the visual encoder and adapter layer, $L$ is the sequence length of answers and $\hat{\mathbf{y}}_{i,<j}$ is all answer tokens before $\hat{\mathbf{y}}_{i,j}$.

\section{Methods}
\label{sec:benchmark}

\begin{figure}[!t]
    \centering
    \includegraphics[width=1.0\linewidth]{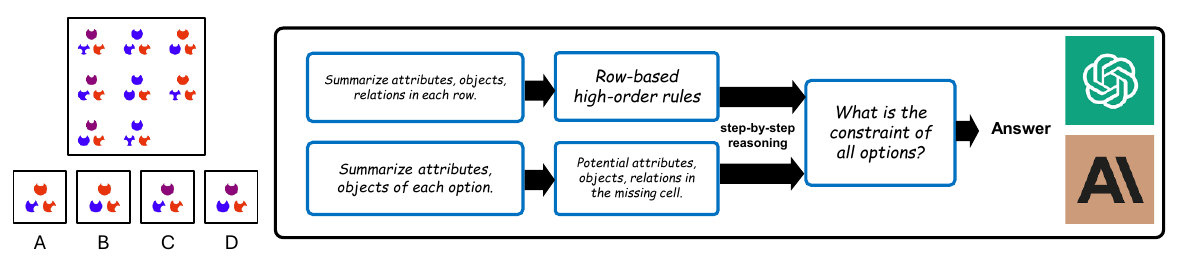}
    \caption{An overview of using CoT to solve matrix reasoning problem in MaRs-VQA. The left part is the model input, including a question image, multiple option images. The right part shows the step-by-step CoT for multi-image reasoning for GPT series and Claude series inference.}
    \label{fig:cot_framework}
\end{figure}

% \begin{figure}[!t]
%     \centering
%     \includegraphics[width=1.0\linewidth]{images/vcogbench.pdf}
%     \caption{An overview of using CoT to solve matrix reasoning problem in MaRs-VQA. The left part is the model input, including a question image, multiple option images and a system prompt describing the task. The right part shows the step-by-step CoT for multi-image reasoning and VLM SFT solution as the baseline method.}
%     \label{fig:baseline}
% \end{figure}

As we claim in previous sections, our initial goal is to complete the $3\times3$ matrix by finding the missing cell from multiple options by \textbf{zero-shot prompt engineering} under the same setting in human's matrix reasoning test. To this end, MLLMs have to deduce relationships across the other cells of the matrix and infer the missing cell accordingly. We use CoT prompt engineering to guide closed-source MLLMs solving this problem. To further promote related visual cognition foundation model, we also propose to use step-by-step structured reasoning annotations in MaRs-VQA to supervised finetune (SFT) Qwen2-VL with LoRA. Then we test the performance of Qwen2-VCog in both MaRs-VQA test set (in-domain) and a subset of RAVEN as Out-of-Domain (OOD) data. The observation in the experiment could reveal why this problem is hard for MLLM and highlight the gap between human intelligence and MLLM reasoning.

\subsection{Multi-Image Reasoning via Chain-of-Thought (CoT)}
\label{sec:method_cot}

\begin{wrapfigure}{R}{0.6\textwidth}
  \vspace{-0.4cm}
  \includegraphics[width=\linewidth]{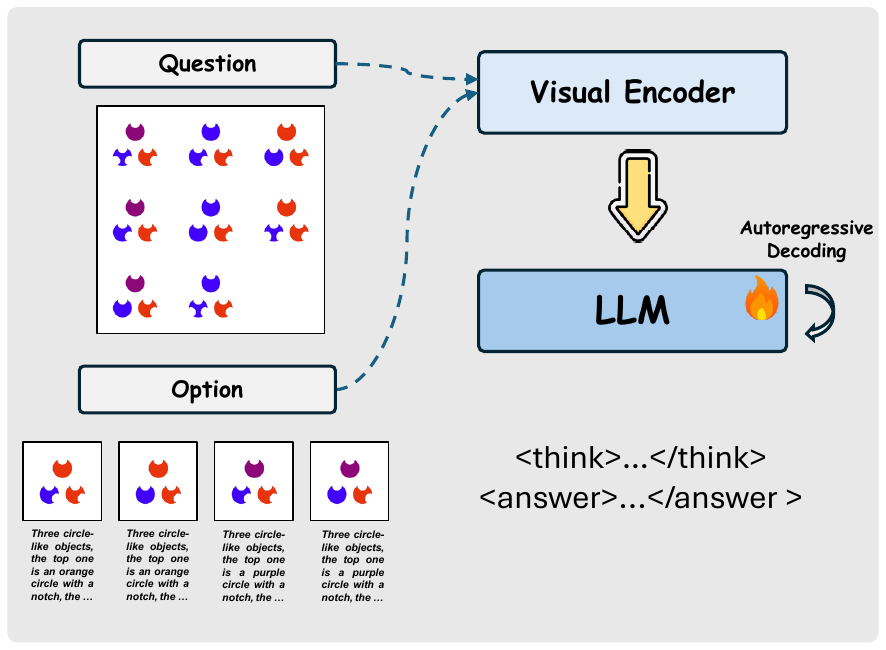}
  \caption{Supervised fine-tuning VLM to generate two-section format for matrix reasoning problem.}
  \label{fig:vlm_framework}
  \vspace{-0.2cm}
\end{wrapfigure}

Building on recent insights into systematic, language-based reasoning~\citep{wei2022emergent,kojima2022large}, we propose a straightforward yet effective division of the reasoning workflow into two distinct and clearly tagged stages (\texttt{<think>} and \texttt{<answer>}), as illustrated in Figure~\ref{fig:vlm_framework}. This design draws inspiration from methods such as OpenAI o1~\citep{zhong2024evaluation}, LLaVA-CoT~\citep{xu2024llava}, and R1-V~\cite{chen2025r1v} where each stage contributes a different level of abstraction to the overall inferential process:
\begin{itemize}
    \item \textbf{Reasoning (<think>)}: It includes: (i) a concise overview of the task (e.g., ‘examining a 3$\times$3 grid puzzle and determining the missing cell’); (ii) a structured description of relevant visual elements (color, shape, position, etc.) that guide the reasoning; and (iii) a methodical analysis of the discovered pattern(s). Crucially, this step covers both rule identification (the model pinpoints how objects or elements follow consistent patterns) and option verification (each candidate option is tested against the identified rule).
    % It includes a concise overview of the task at hand (e.g., examining a 3$\times$3 grid puzzle and determining the missing cell); a structured description of the relevant visual elements, focusing on details (such as color, shape, or position) that guide the reasoning; a methodical analysis of the discovered pattern(s). Crucially, this step covers both (i)~\emph{rule identification} (so the model pinpoints how objects or elements transform or follow consistent patterns) and (ii)~\emph{option verification} (where each candidate option is briefly tested against the identified rule). 
    \item \textbf{Conclusion (<answer>)}: A single, succinct statement that specifies the best or correct choice among the provided options. No extra explanation is given here. It simply states which option is correct as the final answer.
\end{itemize}

This two-section format promotes a structured and transparent reasoning style by first concisely outlining the puzzle, then detailing all pertinent visual characteristics, followed by explicit discovery and testing of candidate rules, and finally isolating the single correct response. To further clarify the CoT processes in LLM-based reasoning, our method annotates each stage of the two-section format with dedicated tags, such as \texttt{<think>...</think>} and \texttt{<answer>...</answer>}. By explicitly marking the beginning and end of each reasoning stage, the model is guided to retain clarity and precision throughout the entire solution path. Unlike the traditional free-form CoT that allows the model to produce unconstrained self-talk, our approach enforces a well-structured methodology. A detailed template demonstrating this two-section CoT format is provided in our code repository.

\subsection{SFT for Vision-Language Model (VLM)}

To enhance the reasoning capabilities of vision-language models (VLMs), we leverage the reasoning responses generated using the two-section CoT format as training data (in Figure~\ref{fig:vlm_framework}). LoRA ~\cite{hu2022lora} is employed to fine-tune Qwen2-VL, enabling efficient adaptation of the model to these structured reasoning tasks. Specifically, the step-by-step reasoning annotations in MaRs-VQA are used as supervision signals during fine-tuning. After training, the performance of the resulting model, Qwen2-VCog, is evaluated on MaRs-VQA test set for in-domain performance and a subset of RAVEN for Out-of-Domain (OOD) performance.

To further probe the contribution of explicit cognitive supervision, we conduct an ablation study by fine-tuning a variant of Qwen2-VL without the step-by-step reasoning annotations, in parallel to Qwen2-VCog (which was trained with full reasoning chains). This allows us to isolate the effect of reasoning supervision on both in-domain (MaRs-VQA) and out-of-domain (RAVEN) performance.

\section{Experiments}
\label{sec:experiments}

\subsection{Experimental Settings} 
\label{sec:exp_setting}

\begin{table}[!t]
  \centering
  \adjustbox{max width=1.0\textwidth}{
  \begin{tabular}{ccccc}
    \toprule
    \multirow{2}{16em}{\textbf{Method}} & \multirow{2}{7em}{Model Scale} & \multicolumn{2}{c}{\textbf{Accuracy (\%) $\uparrow$}} \\
    \cmidrule(lr){3-4}
    && \textbf{MaRs-VQA (4-options)} & \textbf{RAVEN (8-options)}  \\
    \midrule
    \multirow{1}{16em}{Random Select} & - & 25.00 & 12.50 \\
    \multirow{1}{16em}{LLaVA-NExT~\citep{liu2024visual}} & 7B & 16.88 & 14.29 \\
    \multirow{1}{16em}{InternVL-2.5~\cite{chen2024expanding}} & 8B  & 20.00 & 13.21  \\
    \multirow{1}{16em}{Qwen2-VL~\cite{wang2024qwen2}} & 7B   & 23.75 &  29.27 \\
    \multirow{1}{16em}{Claude 3 Sonnet~\citep{claude2024}} & - & 23.22 & 13.39 \\
    \multirow{1}{16em}{Claude 3 Opus~\citep{claude2024}} & -  & 24.13 & 11.95 \\
    \multirow{1}{16em}{Claude 3.5 Sonnet~\citep{claude2024}} & - & 24.28 & 15.36 \\
    \multirow{1}{16em}{GPT-4V~\citep{gpt4v2023}} & - & 33.13 & 15.63 \\
    \multirow{1}{16em}{GPT-4o~\citep{gpt4o2024}} & - & 33.96  & 25.89 \\
    \multirow{1}{16em}{GPT-o1~\citep{gpt4o12024}} & -  & 52.29 & 25.36 \\
    \multirow{1}{16em}{Qwen2-VCog (Our SFT Baseline)} & \multirow{1}*{7B} & 72.71 & 31.96* \\
    \multirow{1}{16em}{Human} & \multirow{1}*{-} & \multirow{1}*{\textbf{81.00}} & \multirow{1}*{\textbf{84.41}} \\
    \bottomrule
  \end{tabular}
  }
  \vspace{-0.1cm}
  \caption{Benchmarking experiments on multi-image reasoning in MaRs-VQA (in-domain) and RAVEN (OOD). zero-shot means only provide the model system prompt about the matrix reasoning task definition. CoT denotes the implementation in section~\ref{sec:method_cot}.
  The results are averaged over three runs with three different random seeds. * of Qwen2-VCog denotes CoT performance without finetuning.}
  \label{tab:cot}
  \vspace{-0.4cm}
\end{table}

\paragraph{Datasets}
For MaRs-VQA, We split 480 VQA samples as the test set and the rest of them is the training set for SFT. In addition to MaRs-VQA, we also select 560 VQA pairs serving as OOD samples from matrix reasoning dataset RAVEN~\citep{zhang2019raven}. Details of difference between MaRs-VQA and RAVEN are shown in Table~\ref{tab:datasets} in appendix.

\paragraph{MLLM Baselines} We selected the Claude 3 Sonnet, Claude 3 Opus, Claude 3.5 Sonnet~\citep{claude2024}, GPT-4V~\citep{gpt4v2023}, GPT-4o~\citep{gpt4o2024}, GPT-o1~\citep{gpto12024}, LLaVA-NExT~\citep{liu2024visual}, Qwen2-VL~\citep{wang2024qwen2}, InternVL-2.5~\cite{chen2024expanding}  as the primary multi-image CoT baselines as they support multiple images input and can generate reasoning process. As Qwen2-VL~\citep{wang2024qwen2} is the best open-sourced VLM for zero-shot inference in Table~\ref{tab:cot}, we choose it as the main backbone to finetune our baseline Qwen2-VCog with MaRs-VQA training data.

\paragraph{Human Baseline}
The human study results in Table~\ref{tab:cot} are reported from previous experiment results. The human subjects of RAVEN~\citep{zhang2019raven} consists of college students from a subject pool maintained by the Department of Psychology. Only ``easily perceptible'' examples were used in the investigation. The human study results of MaRs-IB~\citep{chierchia2019matrix} are more rigorous. They are from 4 age groups ($N = 659$, aged 11–33 years). The accuracy for younger adolescents, mid-adolescents, older adolescents, and adults solving matrix reasoning in MaRs-IB are 61\%, 68\%, 73\%, 81\%. We use the adult result 81\% in Table~\ref{tab:cot}.

\paragraph{Implementation}
For closed-source baseline models, we establish basic prompts to introduce the matrix reasoning problem setting, which serve as the system prompt for zero-shot inference. For open-source baseline models, we use the same system prompt settings across all models. Testing is conducted using two NVIDIA H100 GPUs for all models. All experiments are run with three different random seeds, and the results are averaged. We evaluate the results based on the accuracy of single-option matrix reasoning problems (\( \text{Acc} = \text{Correct} / \text{Total} \)), consistent with other VQA benchmarks~\citep{lu2022learn,liu2023mmbench}.

\subsection{Experimental Results}
\label{sec:experimental_results}

For all models in Table~\ref{tab:cot}, we used multiple images as the input, including a question image and several option images, and guided the MLLMs to decompose the problem into predefined structures before generating answers based on all available information. We tested the latest closed-source models like Claude 3, Claude 3.5, GPT-4V, GPT-4o, and GPT-o1 for this task, as these models can generate step-by-step multi-image reasoning. In addition, we also compare them with small size open-sourced models like Qwen2-VL, InternVL-2.5. Our results show that even the state-of-the-art closed-source MLLMs GPT-o1 perform worse than humans in all matrix reasoning tasks.

After analyzing the reasoning outputs of current MLLMs, we identified three primary issues: (1) Limited Use of Visual Information: MLLMs struggle to directly utilize visual features during reasoning, rendering them insensitive to non-verbal spatial details, particularly evident when interpreting positional relationships within images. For instance, distinguishing among the options in Figure~\ref{fig:example_gpt4v} is challenging for MLLMs using language alone. (2) Restricted Visual Working Memory: MLLMs exhibit limited visual working memory, leading to rapid loss of crucial visual information during text-based reasoning processes. (3) Integration Challenges: Despite excelling at specific visual tasks like recognition, segmentation, and object detection, MLLMs encounter significant difficulties integrating these skills for high-level visual reasoning tasks. Further examples illustrating GPT-4o's failure cases and discussion are provided in the Appendix.

To address these shortcomings, we leveraged the reasoning annotations described in Section~\ref{sec:method_cot} to fine-tune Qwen2-VL using Low-Rank Adaptation (LoRA) as our baseline model for MaRs-VQA. After fine-tuning on the MaRs-VQA training set, the model achieved over 70\% accuracy on the MaRs-VQA test set, nearing human adult performance. Additionally, its performance on the OOD RAVEN dataset improved from 29.27\% to 31.96\%. This observation shows that VLM could exploit shortcut features from matrix reasoning problem. It tends to perform well on the in-distribution test sets while slightly improving the out-of-distribution test performance. This indicates that there is still a major gap for current MLLMs to learn intended features in the abstract reasoning tasks. The simple strategy of SFT cannot entirely solve this task. This also demonstrates the value of our proposed benchmark in examining the out-of-domain nature of abstract reasoning tasks.

\subsection{Ablation Study}
\label{sec:ablation}

\begin{table}[!!t]
    \centering
    \footnotesize
    \begin{minipage}{0.45\columnwidth}
    \centering
    \setlength{\tabcolsep}{2pt} % Adjust column separation
    \adjustbox{max width=1.0\textwidth}{
    \begin{tabular}{lc}
        \toprule
        \textbf{Strategy}  & \textbf{Accuracy (\%)} $\uparrow$ \\
        \midrule
        GPT-4o+ CoT & 33.96 \\
        \midrule
        GPT-4o+ CoT + 1-shot & 35.22 \\
        GPT-4o+ CoT + 3-shot & 36.10 \\
        GPT-4o+ CoT + 5-shot & 36.03 \\
        GPT-4o+ multi-round CoT & 41.96 \\
        GPT-4o+ multi-round CoT + 1-shot & 42.08 \\
        \midrule
        GPT-o1 & 52.29\\
        \bottomrule
    \end{tabular}
    }
    \end{minipage}
    \hfill
    \begin{minipage}{0.50\columnwidth}
    \caption{Ablation on few-shot sample CoT and multi-round CoT for GPT-4o in MaRs-VQA. GPT-o1 still outperforms GPT-4o with different CoT strategy.}
    \label{tab:ablation_prompt}
    \end{minipage}
    \vspace{-0.3cm}
\end{table}

In this subsection, we conduct ablation experiments to analyze how to improve the zero-shot performance of MLLMs on the matrix reasoning problem. 

\paragraph{Effect of CoT Strategy for Closed-source MLLMs}

Table~\ref{tab:ablation_prompt} compares the different CoT strategy: raw step-by-step CoT with hint, CoT reasoning with few-shot sample and multi-round CoT reasoning. Few-shot samples are a small number of question-answer examples alongside the CoT system prompt. Multi-round reasoning employs the advanced multi-round CoT strategy (step by step option elimination). After each round, the model will reflect on the correctness of the answer and run the reasoning steps again if the answer is wrong. The results show that incorporating 1-shot and 3-shot samples gradually increases the accuracy of GPT-4o on MaRs-VQA from 34\% to 36\%. However, extending the number of examples to 5 does not yield further improvement. These findings suggest that while few-shot in-context learning helps the model better understand the matrix reasoning problem, it does not significantly enhance the MLLM's visual reasoning capabilities for these tasks. Additionally, using a multi-round elimination strategy improves accuracy from approximately 34\% to 42\%, but it is considerably slower than single-round CoT, and still cannot surpass GPT-o1 (52.29\%) and human adults (81.00\%).

% To further analyze the sources of MLLM limitations and the effect of our methodology, we conducted three additional experiments, detailed in the Appendix: (1)~An ablation on the effect of explicit reasoning strings during SFT, which demonstrates the necessity of step-by-step cognitive supervision for robust performance; (2)~A targeted comparison probing whether vision or language constitutes the primary bottleneck in matrix reasoning, showing that visual pattern extraction is the main limiting factor; and (3)~A comprehensive evaluation on general VLM benchmarks to ensure that our reasoning-focused SFT does not degrade overall multimodal language capabilities. Please refer to Appendix~\ref{app:reasoning_sft}, \ref{app:vision_language}, and \ref{app:general_bench} for details and results. For further analysis of shortcut learning and domain shift between MaRs-VQA and RAVEN, see Appendix~\ref{app:shortcut}.

\paragraph{Effect of Reasoning Strings in SFT}
\label{app:reasoning_sft}

To more precisely quantify the impact of explicit step-by-step reasoning supervision in Supervised Fine-Tuning (SFT), we conducted an ablation study in which the reasoning strings were omitted from the MaRs-VQA training data. As reported in Table~\ref{tab:reasoning_sft}, the absence of these structured cognitive annotations led to a dramatic decrease in accuracy for Qwen2-VCog on both MaRs-VQA and RAVEN. Specifically, performance on MaRs-VQA dropped by nearly 18 percentage points, while the accuracy on RAVEN also decreased, albeit to a lesser extent. This result highlights that detailed reasoning guidance is not merely auxiliary, but instead serves as a crucial signal for the model to acquire transferable visual cognitive skills. Without such supervision, the model tends to rely on shallow pattern matching or spurious correlations, failing to generalize robustly even within the same task domain. These findings underscore the necessity of incorporating high-quality, stepwise annotations for training MLLMs to approach human-like reasoning on complex visual cognition benchmarks.

\begin{table}[h]
\scriptsize
\centering
\adjustbox{width=0.7\textwidth}{
\begin{tabular}{lcc}
\toprule
\textbf{Model} & \textbf{MaRs-VQA (\%)} $\uparrow$ & \textbf{RAVEN (\%)} $\uparrow$ \\
\midrule
Qwen2-VCog (with reasoning) & 72.71 & 31.96 \\
Qwen2-VCog (w/o reasoning) & 54.82 & 29.30 \\
\bottomrule
\end{tabular}
}

\caption{Effect of reasoning strings in SFT.}
\label{tab:reasoning_sft}

\end{table}

\paragraph{Vision vs. Language Bottleneck}
\label{app:vision_language}

We further investigated whether the primary limitation of current MLLMs in matrix reasoning tasks arises from their visual perception modules or from downstream language-based reasoning. To disentangle these factors, we designed three input conditions: (1) providing only the question image and selecting options via CLIP similarity, (2) providing both the question and all option images (the standard VQA setting), and (3) supplementing the images with perfect human-annotated textual descriptions for both the question and options. As shown in Table~\ref{tab:vision_language}, both GPT-4o and Qwen2-VCog perform at near chance level when deprived of option images, indicating a failure to extract the underlying visual rules from the question image alone. When option images are provided, accuracy increases substantially, especially for Qwen2-VCog, which demonstrates strong visual pattern matching. Notably, the addition of ideal textual descriptions yields only a marginal further improvement, suggesting that the language reasoning component is not the main bottleneck once high-quality visual features are available. These results collectively point to visual pattern extraction—rather than linguistic inference—as the principal limiting factor for MLLMs on abstract visual reasoning tasks, emphasizing the need for stronger visual encoders and better integration of visual working memory.

\begin{table}[h]
\scriptsize
\centering
\adjustbox{width=0.9\textwidth}{
\begin{tabular}{lcc}
\toprule
\textbf{Input Setting} & \textbf{GPT-4o (\%)} $\uparrow$ & \textbf{Qwen2-VCog (\%)} $\uparrow$ \\
\midrule
Question image only & 24.58 & 26.32 \\
Question image + Option images & 33.96 & 72.71 \\
Question image + Option images + Option Description & 36.46 & 71.43 \\
\bottomrule
\end{tabular}
}

\caption{Vision vs. Language Bottleneck Analysis.}
\label{tab:vision_language}

\end{table}

\paragraph{Generalizability}
\label{app:general_bench}

To ensure that the improvements observed from SFT on MaRs-VQA do not come at the expense of general visual-language understanding, we systematically evaluated both Qwen2-VCog and the original Qwen2-VL on a suite of standard multimodal benchmarks, including MME~\citep{fu2024mmecomprehensiveevaluationbenchmark}, HallusionBench~\citep{guan2024hallusionbench}, POPE~\citep{li2023evaluating}, VQAv2 (validation set)~\citep{goyal2017making}, SQA\_Image~\citep{lu2022learn}, and SeedBench~\citep{li2024seed}. As summarized in Table~\ref{tab:general_bench}, the two models exhibit nearly identical performance across all tasks, with only negligible fluctuations that are well within the range of experimental noise. This demonstrates that reasoning-focused finetuning on MaRs-VQA does not degrade the model's ability to perform generic vision-language tasks, nor does it induce catastrophic forgetting. Consequently, it is feasible to endow MLLMs with enhanced visual reasoning skills via SFT on MaRs-VQA, without sacrificing their broader applicability or real-world utility.

\begin{table}[h]
\scriptsize
\centering
\adjustbox{width=\textwidth}{
\begin{tabular}{lcccccc}
\toprule
\textbf{Model} & \textbf{MME} $\uparrow$ & \textbf{Hallusion\_Bench} $\uparrow$ & \textbf{POPE} $\uparrow$ & \textbf{VQAv2\_Val} $\uparrow$ & \textbf{SQA\_Image} $\uparrow$ & \textbf{SeedBench} $\uparrow$ \\
\midrule
Qwen2-VL-7B & 1666 & 53.83 & 88.84 & 79.88 & 83.73 & 69.16 \\
Qwen2-VCog-7B & 1680 & 55.73 & 88.71 & 79.45 & 83.29 & 68.87 \\
\bottomrule
\end{tabular}
}

\caption{General VLM Benchmarks: SFT does not degrade generic ability.}
\label{tab:general_bench}

\end{table}

\subsection{Visualization}

\begin{figure}[!t]
    \centering
    \includegraphics[width=1.0\linewidth]{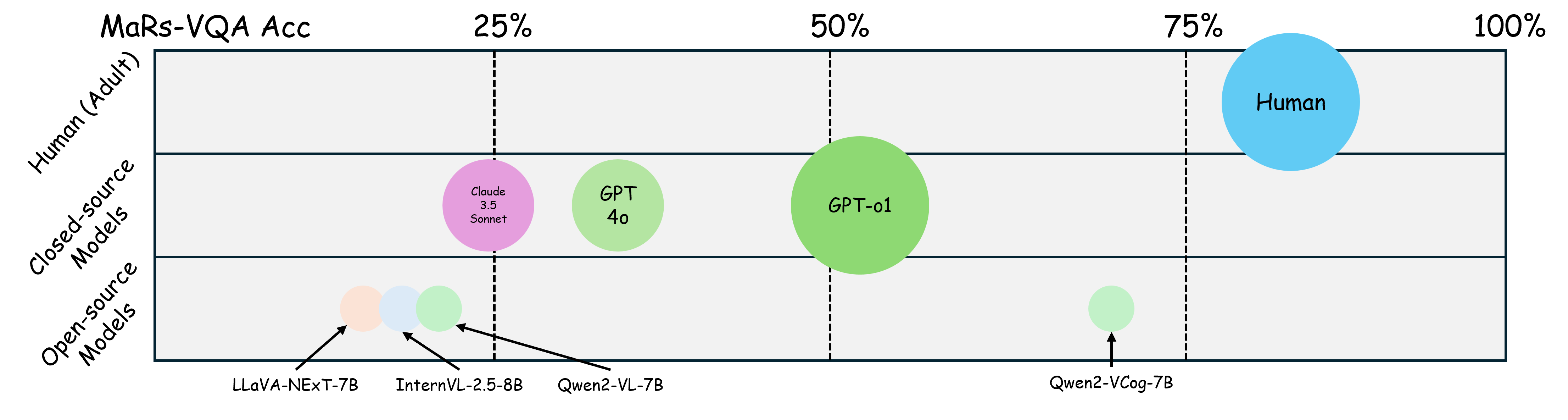}
    \caption{There is still a substantial gap between MLLM's (zero-shot CoT or SFT training) matrix reasoning capability and human's (zero-shot). Bubble size corresponds to the model size. As we don't know the exact size of closed-source MLLMs, we set all of them to the largest value by default. The model size of human refers to the number of neurons (86B) in human's brain~\citep{voytek2013there}.}
    \label{fig:accuracy_size_vis}
\end{figure}

We also analyze the relationship between matrix reasoning accuracy and model scale in Figure~\ref{fig:accuracy_size_vis}. The figure illustrates the significant gap between MLLM's matrix reasoning performance and that of humans. This gap is substantial and suggests that simply increasing model size according to scaling laws will not be sufficient to bridge it. 

% \begin{figure}[h]
%     \centering
%     \footnotesize
%     \begin{minipage}{0.6\columnwidth}
%     \centering
%     \includegraphics[width=\linewidth]{images/accuracy_size.png}
%     \end{minipage}
%     \hfill
%     \begin{minipage}{0.37\columnwidth}
%     \caption{There is still a big gap between human's (zero-shot) matrix reasoning capability and MLLM's. Bubble size corresponds to the model size. As we don't know the exact size of closed-source MLLMs, we set all of them to the largest value by default. The model size of human refers to the number of neurons (86B) in human's brain~\citep{voytek2013there}.}
%     \label{fig:accuracy_size_vis}
%     \end{minipage}
%     \vspace{-0.3cm}
% \end{figure}

\section{Discussion}
\label{sec:discussion}

In this work, we emphasize that zero-shot matrix reasoning is a crucial testbed for human-level intelligence, even though the developmental origins of this ability in children remain unclear. Remarkably, children as young as four can solve matrix reasoning problems without explicit training, highlighting the unique strengths of human visual cognition. Our long-term goals are twofold: (1) to rigorously evaluate how close MLLMs are to human-like cognitive abilities, as posed by \citet{chollet2019measure}; and (2) to develop MLLM-powered agents capable of human-level zero-shot matrix reasoning, which could in turn generate novel assessment tools to help psychologists and pediatricians understand neurodevelopment.

Our findings reveal a persistent visual cognition gap between current MLLMs and humans, even as model scale increases. Detailed ablation and analysis suggest that the primary bottleneck lies in visual pattern extraction and working memory, rather than language reasoning. This gap has concrete implications for real-world applications: it limits the reliability of AI in dynamic or safety-critical environments (e.g., robotics, scientific discovery, education), where robust abstract visual reasoning and generalization are essential. Bridging this gap will require advances in both data and architecture, particularly in strengthening the visual encoders and multimodal integration of MLLMs. Our benchmark and insights lay the groundwork for future research toward more human-like, generalizable visual cognition in AI systems.

\section{Conclusion}
\label{sec:conclusion}

We introduce MaRs-VQA, a publicly available matrix reasoning Visual Question Answering (VQA) dataset specifically designed to evaluate the visual cognitive capabilities of Multimodal Large Language Models (MLLMs) and compare them with humans. Our findings indicate that state-of-the-art MLLMs, such as GPT-4o, Qwen2-VL, and InternVL-2.5, demonstrate foundational competence in matrix reasoning but continue to struggle with more complex or abstract scenarios, performing substantially below human levels. Supervised fine-tuning (SFT) with cognitively designed step-by-step reasoning annotations from MaRs-VQA can significantly boost in-domain accuracy, yet these models still fall short of human performance and generalize poorly to out-of-domains (OOD). Notably, humans achieve strong performance without any task-specific training, underscoring the inherent gap in visual cognition between humans and MLLMs. Our ablations further show that explicit reasoning supervision is crucial, and that vision—not language—remains the dominant bottleneck. Bridging this gap will require continued research and innovation in both model architecture and multimodal learning paradigms, ultimately advancing the visual cognitive abilities of future MLLMs.
\clearpage
\bibliography{colm2025_conference}
\bibliographystyle{colm2025_conference}

% \appendix
\clearpage
\appendixpage

\appendix

\tableofcontents

\clearpage

\section{Data Collection and Licenses}

We showed and compared MaRs-VQA and RAVEN in Table~\ref{tab:datasets}. The reason we choose RAVEN, MaRs-VQA is because all these datasets contain zero-shot / few-shot human investigation results in their follow-up studies. Based on these results, we can compare the MLLM's performance with human in matrix reasoning tasks.

For RAVEN, we followed the original data generation pipeline in their repo. For MaRs-VQA, we download all questionnaires from MaRs-IB and then re-annotate all images by ourselves.

\paragraph{RAVEN}
The original dataset link of RAVEN is \href{https://github.com/WellyZhang/RAVEN}{github.com/WellyZhang/RAVEN}. It is under GPL-3.0 License (\href{https://github.com/WellyZhang/RAVEN/blob/master/LICENSE}{RAVEN LICENSE}) and is free to use by public. All data in RAVEN are generated by rule-based scripts. We follow the basic setting of RAVEN, and modify the range of $\text{COLOR\_VALUES}$ to $[255, 192, 128, 64, 0]$ and $\text{SIZE\_VALUES}$ to $[0.3, 0.45, 0.6, 0.75, 0.9]$. The sample size of RAVEN is 560.

\paragraph{MaRs-VQA}

The image data of MaRs-VQA is from MaRs-IB~\citep{chierchia2019matrix} and annotated with context option by our team. It contains 18 questionnaires, each of questionnaire contains 80 matrix reasoning questions. The human study of MaRs-IB is rigorous. In MaRs-IB's original user study, all participants provided informed consent and all procedures were approved by UCL's ethical committee. 

The paper and study results are under MIT License. All questionnaires are under Attribution-NonCommercial 3.0 (\href{https://osf.io/gkvy4}{MaRs-IB LICENSE}), which means it allows people to use the work, or adaptations of the work, for noncommercial purposes only, and only as long as they give credit to the creator. Thus, the MaRs-VQA dataset will under the same license.

The sub-task statistics of MaRs-VQA is in Table.

Compared to other zero-shot matrix reasoning dataset (Table~\ref{tab:dataset_comparison}) to evaluate matrix reasoning for MLLMs, MaRs-VQA has advantages list below:

\begin{itemize}
    \item MaRs-VQA comprises 1,440 image instances designed by psychologists, making it the largest dataset for zero-shot matrix reasoning evaluation.

    \item MaRs-VQA includes a diverse range of data, such as variations in color, geometry, positional relationships, and counting.

    \item The data source for MaRs-VQA is MaRs-IB~\citep{chierchia2019matrix}, which is based on rigorous human studies. This dataset is widely recognized in the psychology community and has inspired numerous follow-up studies in child psychology and pediatrics. This is the first time we introduce it to the AI/ML community.
\end{itemize}

\section{MaRs-VQA Difficulty Level Study}
\label{sec:diff_study}

We also compare GPT-4o across difficulty levels and different visual complexities in the MaRs-VQA dataset in Table~\ref{tab:difficulty}. In Table~\ref{tab:marsvqa_difficulty}, the difficulty of matrix reasoning tasks can be categorized into five levels based on the complexity of attribute changes: Difficulty Level 1 involves a single changing attribute (e.g., shape, color, size, position, or multi-object) or two simple attributes; Difficulty Level 2 combines multi-object attributes with one other attribute (e.g., shape, color, size, or position); Difficulty Level 3 involves three simultaneously changing simple attributes (e.g., shape, color, and size); Difficulty Level 4 combines multi-object attributes with two other attributes (e.g., shape and color); and Difficulty Level 5 and above includes combinations of four or more attributes. The difficulty increases as the number and complexity of attribute combinations grow. The results indicate that GPT-4o exhibits difficulty sensitivity similar to that of humans. This is because GPT-4o can solve object size sub-tasks well in the MaRs-VQA, but is still struggling with other sub-tasks, especially the multi-object sub-task.

\begin{table}[h]\scriptsize
  \centering
  \begin{tabular}{ c  c  c  c}
    \toprule
    \textbf{Difficulty Level} & \textbf{Question} & \textbf{Option} & \textbf{Description} \\
    \midrule
    \textbf{Level 1} & \begin{minipage}[b]{0.2\columnwidth}
		\centering
		\raisebox{-.45\height}{\includegraphics[width=0.6\linewidth]{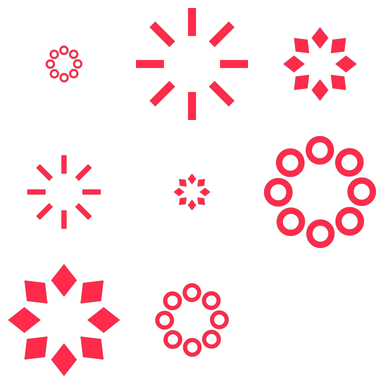}}
	\end{minipage} & \begin{minipage}[b]{0.2\columnwidth}
		\centering
		\raisebox{-.5\height}{\includegraphics[width=0.2\linewidth]{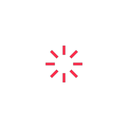}}
      \raisebox{-.5\height}{\includegraphics[width=0.2\linewidth]{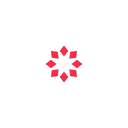}}
      \raisebox{-.5\height}{\includegraphics[width=0.2\linewidth]{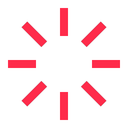}}
      \raisebox{-.5\height}{\includegraphics[width=0.2\linewidth]{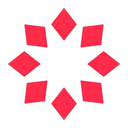}}
	\end{minipage} & Shape + Size \\
    \midrule
    \textbf{Level 2} & \begin{minipage}[b]{0.2\columnwidth}
		\centering
		\raisebox{-.45\height}{\includegraphics[width=0.6\linewidth]{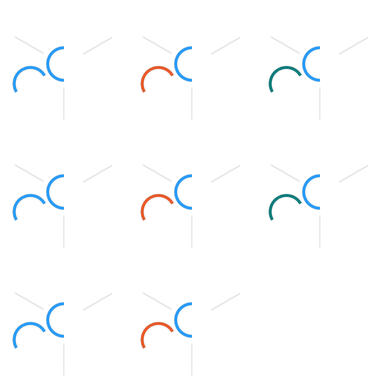}}
	\end{minipage} & \begin{minipage}[b]{0.2\columnwidth}
		\centering
		\raisebox{-.5\height}{\includegraphics[width=0.2\linewidth]{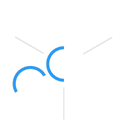}}
      \raisebox{-.5\height}{\includegraphics[width=0.2\linewidth]{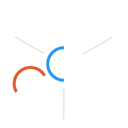}}
      \raisebox{-.5\height}{\includegraphics[width=0.2\linewidth]{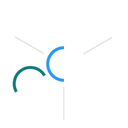}}
      \raisebox{-.5\height}{\includegraphics[width=0.2\linewidth]{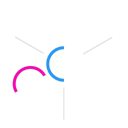}}
	\end{minipage} & Color + Multi-object \\
    \midrule
    \textbf{Level 3} & \begin{minipage}[b]{0.2\columnwidth}
		\centering
		\raisebox{-.45\height}{\includegraphics[width=0.6\linewidth]{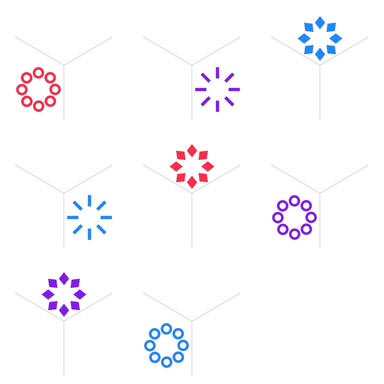}}
	\end{minipage} & \begin{minipage}[b]{0.2\columnwidth}
		\centering
		\raisebox{-.5\height}{\includegraphics[width=0.2\linewidth]{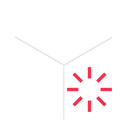}}
      \raisebox{-.5\height}{\includegraphics[width=0.2\linewidth]{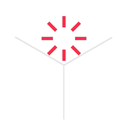}}
      \raisebox{-.5\height}{\includegraphics[width=0.2\linewidth]{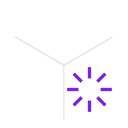}}
      \raisebox{-.5\height}{\includegraphics[width=0.2\linewidth]{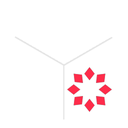}}
	\end{minipage} & Shape + Color + Position \\
    \midrule
    \textbf{Level 4} & \begin{minipage}[b]{0.2\columnwidth}
		\centering
		\raisebox{-.45\height}{\includegraphics[width=0.6\linewidth]{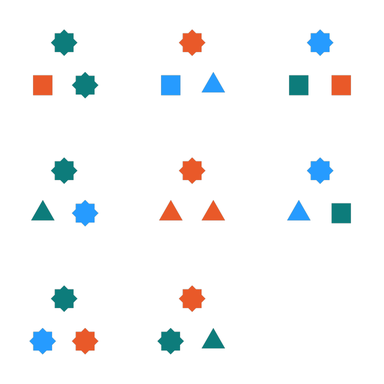}}
	\end{minipage} & \begin{minipage}[b]{0.2\columnwidth}
		\centering
		\raisebox{-.5\height}{\includegraphics[width=0.2\linewidth]{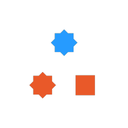}}
      \raisebox{-.5\height}{\includegraphics[width=0.2\linewidth]{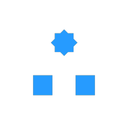}}
      \raisebox{-.5\height}{\includegraphics[width=0.2\linewidth]{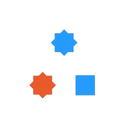}}
      \raisebox{-.5\height}{\includegraphics[width=0.2\linewidth]{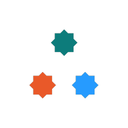}}
	\end{minipage} & Shape + Color + Multi-object \\
    \midrule
    \textbf{Level 5} & \begin{minipage}[b]{0.2\columnwidth}
		\centering
		\raisebox{-.45\height}{\includegraphics[width=0.6\linewidth]{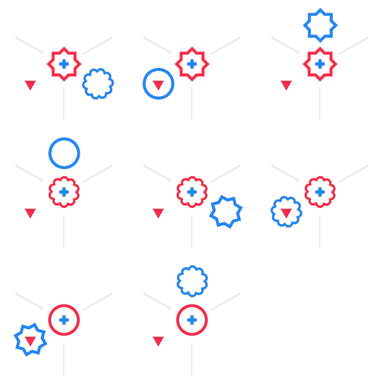}}
	\end{minipage} & \begin{minipage}[b]{0.2\columnwidth}
		\centering
		\raisebox{-.5\height}{\includegraphics[width=0.2\linewidth]{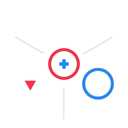}}
      \raisebox{-.5\height}{\includegraphics[width=0.2\linewidth]{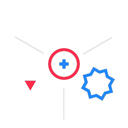}}
      \raisebox{-.5\height}{\includegraphics[width=0.2\linewidth]{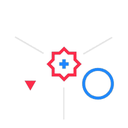}}
      \raisebox{-.5\height}{\includegraphics[width=0.2\linewidth]{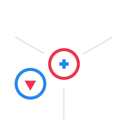}}
	\end{minipage} & Shape + Color + Position + Multi-object \\
    \bottomrule
  \end{tabular}
  \vskip .5 em
  \caption{Explanation of Difficulty Levels in MaRs-VQA.}
  \label{tab:marsvqa_difficulty}
  \vspace{-0.3cm}
\end{table}

\begin{table}[h]\scriptsize
  \centering
  \adjustbox{width=0.9\textwidth}{
  \begin{tabular}{ccccccc}
    \toprule
    \multirow{2}*{\textbf{Method}} &\multicolumn{5}{c}{\textbf{Accuracy (\%)} $\uparrow$} \\
    \cmidrule(lr){2-6}
    &  \textbf{Level 1 (90)} & \textbf{Level 2 (96)} & \textbf{Level 3 (84)} & \textbf{Level 4 (72)} & \textbf{Level 5 (138)}
     \\
    \midrule
    GPT-4o + CoT & 57.78 & 27.08 & 27.38 & 19.43 & 21.74\\
    \bottomrule
  \end{tabular}
  }
  \vspace{-0.1cm}
  \caption{Test GPT-4o with different difficulty levels in MaRs-VQA. The number in the ``()'' is the number of case sample of selected level. The difficulty level is based on the complexity of color, size, geometry, positional relationships, and object counting (See Appendix for more details).}
  \label{tab:difficulty}
  \vspace{-0.5cm}
\end{table}

\section{Experimental Settings}
\label{appendix:experiment}

\begin{table}[!ht]\scriptsize
  \centering
  \begin{tabular}{ c  c  c  c  c}
    \toprule
    \textbf{Dataset} & \textbf{Question} & \textbf{Option} & \textbf{Instance} & \textbf{Description} \\
    \midrule
    \thead{\scriptsize RAVEN \\ \scriptsize \citep{zhang2019raven}} &
    \begin{minipage}[b]{0.2\columnwidth}
		\centering
		\raisebox{-.5\height}{\includegraphics[width=0.6\linewidth]{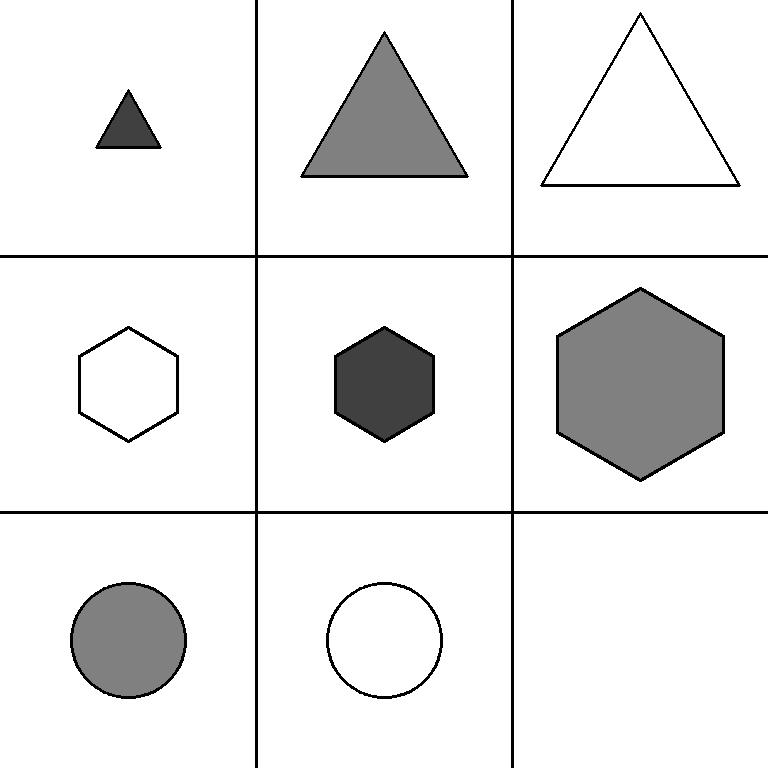}}
	\end{minipage}
    & \begin{minipage}[b]{0.2\columnwidth}
		\centering
		\raisebox{-.95\height}{\includegraphics[width=0.2\linewidth]{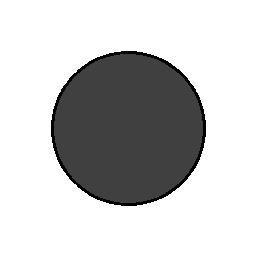}}
      \raisebox{-.95\height}{\includegraphics[width=0.2\linewidth]{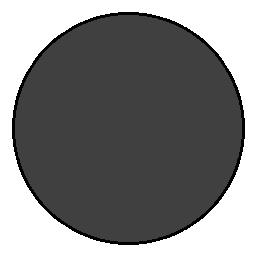}}
      \raisebox{-.95\height}{\includegraphics[width=0.2\linewidth]{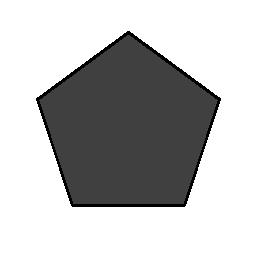}}
      \raisebox{-.95\height}{\includegraphics[width=0.2\linewidth]{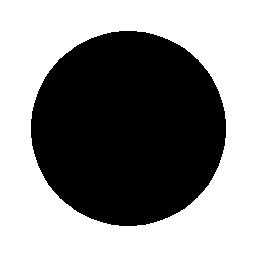}} \\
      		\raisebox{-.95\height}{\includegraphics[width=0.2\linewidth]{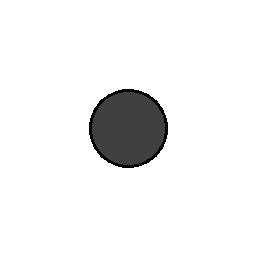}}
      \raisebox{-.95\height}{\includegraphics[width=0.2\linewidth]{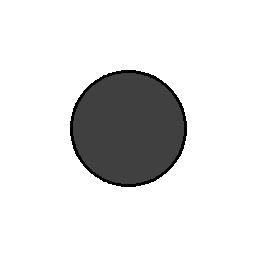}}
      \raisebox{-.95\height}{\includegraphics[width=0.2\linewidth]{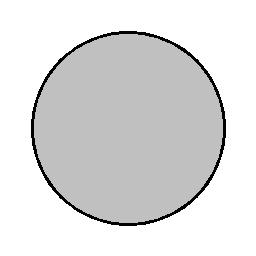}}
      \raisebox{-.95\height}{\includegraphics[width=0.2\linewidth]{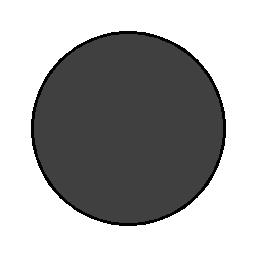}}
	\end{minipage}
    & \thead{\scriptsize rule-based \\ \scriptsize generation}
    & \thead{\scriptsize 8 options per instance \\ \scriptsize grayscale image \\ \scriptsize rule-based stimuli \\ \textbf{\scriptsize include human study}}
    \\ 
    \midrule
    MaRs-VQA &
    \begin{minipage}[b]{0.2\columnwidth}
		\centering
		\raisebox{-.45\height}{\includegraphics[width=0.6\linewidth]{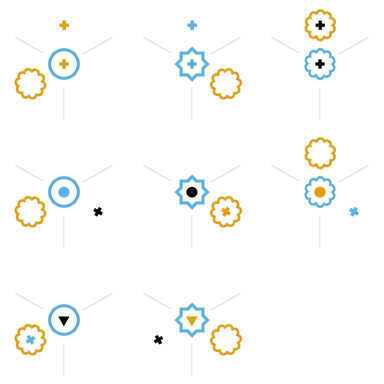}}
	\end{minipage}
    & \begin{minipage}[b]{0.2\columnwidth}
		\centering
		\raisebox{-.5\height}{\includegraphics[width=0.2\linewidth]{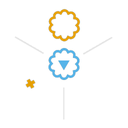}}
      \raisebox{-.5\height}{\includegraphics[width=0.2\linewidth]{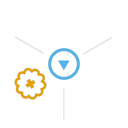}}
      \raisebox{-.5\height}{\includegraphics[width=0.2\linewidth]{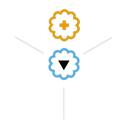}}
      \raisebox{-.5\height}{\includegraphics[width=0.2\linewidth]{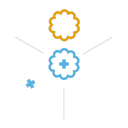}}
	\end{minipage}
    & 1,440
    & \thead{\scriptsize 4 options per instance \\ \scriptsize RGB image \\ \scriptsize psychologist designed stimuli \\ \textbf{\scriptsize include human study}}
    \\ 
    \bottomrule
  \end{tabular}
  \vskip .5 em
  \caption{Experiment datasets comparison. The OOD dataset RAVEN is rule-based generated datasets. The test samples in MaRs-VQA are designed by psychologists from MaRs-IB.}
  \label{tab:datasets}
\end{table}

\subsection{Implementation Details}

We used langchain to implement all closed-source MLLMs. The temperature of all models are 0 and the max token length is 1024. For all datasets, we follow their default image size, type settings for closed-source MLLMs. All experiments are run with three different random seeds, however, since we set temperature to 0, the final accuracy is the same for all random seeds. 

For open-source models, we use the public available weights and data loader settings from the HuggingFace. Testing is conducted using two NVIDIA H100 GPUs for all VLMs. All experiments are run with three different random seeds, and the results are averaged.

\begin{figure}[!t]
    \centering
    \includegraphics[width=1.0\linewidth]{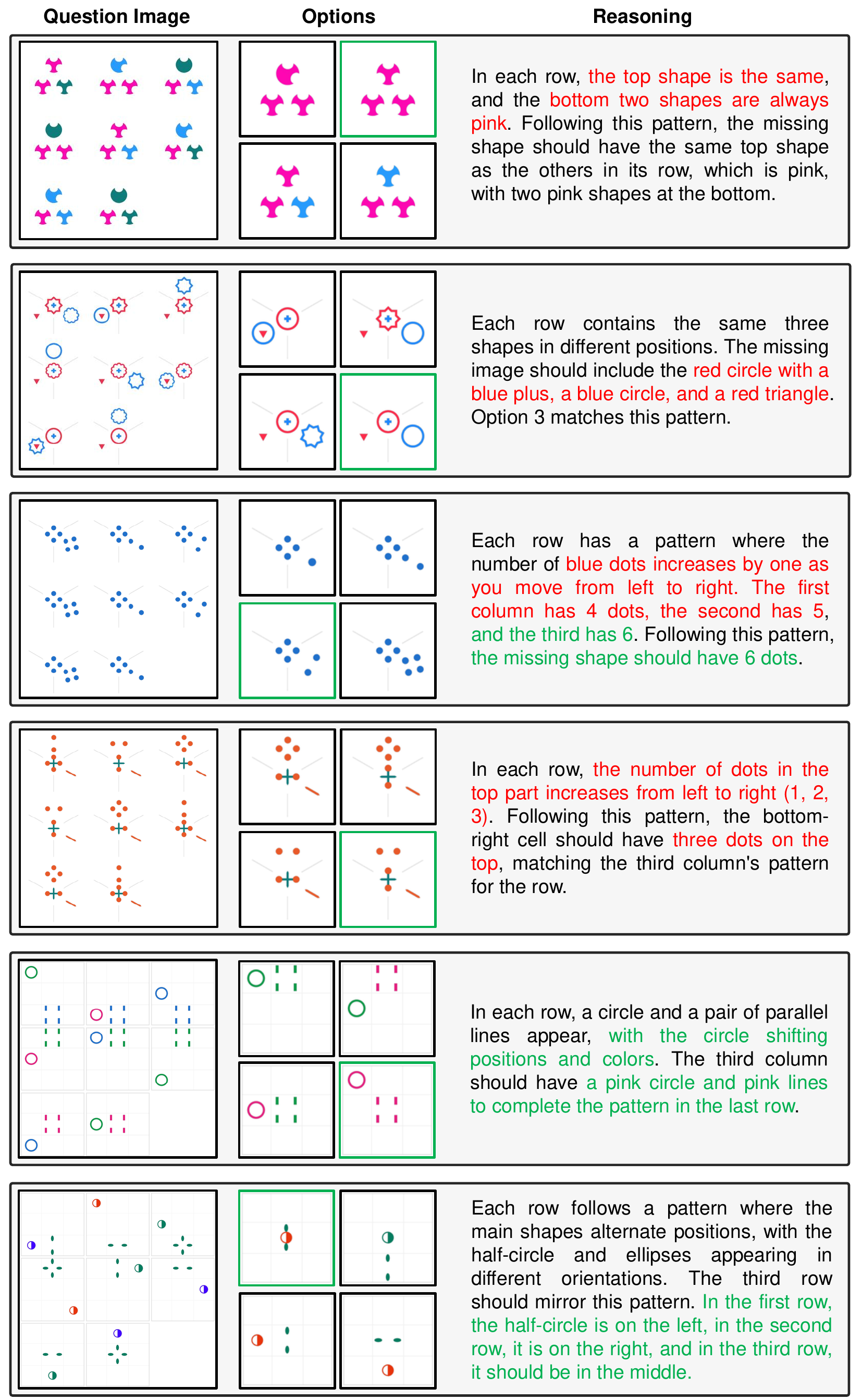}
    \caption{More visualization results for GPT-4o's reasoning.}
    \label{fig:visualization_appendix}
\end{figure}

Based on Figure~\ref{tab:marsvqa_difficulty}, here is the explanation of difficulty levels presented in our paper:

\begin{itemize}
\item \textbf{Difficulty Level 1}: Single sub-task and two simple sub-tasks
Description: The task involves only one changing attribute across the matrix reasoning—either shape, color, size, position, or multi-object. Or two simple attributes: (shape \& color), (shape \& size), (shape \& position), (color \& size), (color \& position), (size \& position).
Example: Figure 4 (top-left) is a matrix reasoning task where only the size and color of the objects changes. This is a difficulty level 1 task.

\item \textbf{Difficulty Level 2}: Two sub-tasks involving multi-object sub-task
Description: The task involves multiple objects combined with one other changing attribute. The sub-task combinations are (multi-object \& shape), (multi-object \& color), (multi-object \& size), (multi-object \& position).

\item \textbf{Difficulty Level 3}: Three simple sub-tasks combined
Description: The task involves three changing attributes simultaneously. The sub-task combinations are (shape \& color \& size), (shape \& position \& size), (shape \& position \& color), (size \& position \& color).

\item \textbf{Difficulty Level 4}: Three sub-tasks involving multi-object sub-task
Description: The task involves multiple objects combined with two other changing attributes. The sub-task combinations are (multi-object \& shape \& color), (multi-object \& shape \& size), (multi-object \& shape \& position), (multi-object \& color \& position), (multi-object \& color \& size), (multi-object \& position \& size).

\item \textbf{Difficulty Level 5 and Above}: Four or more Sub-tasks
Description: The task involves combinations of four or five attributes.
Example: Figure 4 (top-right) is a matrix reasoning task (shape \& position \& color \& multi-objects) and its difficulty level is > 4.
\end{itemize}

As more attributes change simultaneously, the task becomes more complex, requiring higher levels of abstract reasoning to identify patterns. In addition, each additional changing element adds to the cognitive load, making it more challenging to discern the correct answer.

\subsection{More Qualitative analysis}

In this section, we further analyze the failure cases of GPT-4o. Correct reasoning is highlighted in green, while incorrect reasoning is marked in red. Although GPT-4o is sometimes able to extract a subset of key information from the question image, it frequently fails to arrive at the correct final answer. This is primarily due to critical features being either overlooked or inadequately utilized in the decision-making process. As a result, the final answers are often incorrect or only partially aligned with the relevant attributes. It reveals that visual working memory will be a key part to optimize the MLLM's performance in matrix reasoning problem.

\section{Further Discussion on Limitations}

\paragraph{Insights}

Unlike other VQA benchmarks, our work approaches the perspective of human visual cognition—an underexplored domain. Based on our experimental results, we offer the following insights for vision researchers:

\begin{itemize} 
\item While scaling laws have some applicability to visual cognition tasks, merely increasing model size and training data is insufficient to achieve human-level performance. 
\item To demonstrate that VLMs possess strong visual cognitive abilities, it is crucial to evaluate them on zero-shot inference tasks like matrix reasoning—tasks characterized by simple visual content but requiring complex reasoning to find the correct answer. 
\item Unlike other multi-image visual reasoning benchmarks, MaRs-VQA effectively highlights the performance gap between MLLMs and human cognition in these tasks. 
\end{itemize}

From our main and ablation experiments, we observed that as task difficulty increases, the performance of MLLMs in multi-image reasoning scenarios deteriorates. Interestingly, providing language-based descriptions of each option (i.e., inputting the model with a single question image and context-based options) improved the models' performance compared to using multi-image options. This suggests that language still plays a significant role in the visual reasoning processes of current MLLMs and VLMs.

In contrast, human visual cognition—especially in children—allows individuals to solve matrix reasoning tasks without relying on advanced language reasoning capabilities. Children can often solve these tasks effectively by utilizing their visual working memory and pattern recognition skills.

One potential reason for the performance gap is that current MLLMs/VLMs may underemphasize the visual encoder relative to the language encoder. In many recently released VLMs, the visual module is much smaller than the language model module, and the visual encoders are frozen during Large Language Model (LLM) and alignment layer fine-tuning in open-sourced VLMs. This imbalance might limit the models' capacity to retain and process complex visual information during reasoning tasks.

To better retain visual information during the reasoning process, MLLMs may require more capable visual modules that can handle complex visual patterns and maintain this information throughout the reasoning steps. Moreover, optimizing the training process with end-to-end multimodal training—without freezing any layers in the visual modules—can be beneficial. Recent models have begun to explore end-to-end VLM fine-tuning, demonstrating the potential of this approach, though challenges remain such as the need for multi-round alignment. In the future, developing more advanced methods to effectively integrate visual and linguistic features will be crucial.

\paragraph{Limitations}

In the main paper, we briefly discussed the limitations of our work. Here, we provide a more in-depth discussion. First, our dataset is composed of limited publicly available matrix reasoning datasets, which must include human study results. The RAVEN, created by the AI/ML community, were not developed following rigorous psychological research norms. Consequently, our benchmarking results, which utilize these datasets, should not be used to derive psychological or clinical conclusions. While MaRs-VQA addresses this problem, its samples cannot represent all formats of matrix reasoning found in IQ tests such as the WISC and the Cattell Culture Fair Intelligence Test~\citep{cattell1960measuring}. We cannot use these IQ tests directly because they are not freely available, and copyright restrictions usually prevent these pen-and-paper tasks from being adapted into computerized formats.

Second, the size of MaRs-VQA is relatively small compared with typical computer vision datasets, due to the inherent challenges involved in collecting matrix reasoning data. However, as we have argued in our paper, matrix reasoning should not be presented in typical machine learning settings—fine-tuning models on training sets and evaluating performance on test sets. Benchmarking MLLMs' visual reasoning performance should be conducted in a zero-shot inference setting, ensuring that all data in the test set are not included in the models' training data. Even compared with other recently released human-designed matrix reasoning datasets, ours is still the largest (see Table~\ref{tab:dataset_comparison}).

\paragraph{Future Work}

% Although LLMs have achieved remarkable success in language understanding and generation, a significant portion of their parameters is dedicated to encoding linguistic patterns and memorizing factual information, which offers limited benefits for tasks requiring visual cognition. This disparity between Multimodal LLMs and humans indicates that merely increasing model size is insufficient to achieve human-level zero-shot inference in these domains. While our benchmark and baseline models represent a significant initial step, further data collection and in-depth human studies remain essential.

% From our experimental results, we observe that current MLLMs have enhanced basic matrix reasoning capabilities, with models like GPT-4o and Gemini Pro 1.5 achieving significantly higher accuracy than random guessing across all three matrix reasoning tasks. By using Monte Carlo Tree Search to optimize the results via multi-round reasoning and exclusion, GPT-4o can achieve much better outcomes, albeit at the cost of increased inference time. We anticipate that the next generation of MLLMs will approach human-level performance in matrix reasoning. It is crucial to maintain these visual cognition-based benchmarks, continuously monitor the performance of newly released MLLMs, and encourage open-source MLLMs and VLMs to include matrix reasoning tasks for performance comparison.

Finally, we pose the open-ended question of whether MLLMs need to achieve or surpass human-level zero-shot inference capability in matrix reasoning tasks. Addressing this issue requires drawing on theories from cognitive science and psychology to understand the nature of human and MLLM intelligence. Matrix reasoning ability develops early in human neurodevelopment, with children as young as four providing sensible answers to simple matrix reasoning questions without additional training, making it a critical component of IQ tests. In contrast, LLMs and MLLMs rely on training data, fundamentally differing from how children develop cognitive abilities. However, we believe that these two learning processes share commonalities: both involve the gradual accumulation of skills and the ability to generalize from past experiences. Exploring these parallels can provide valuable insights into designing MLLMs that more closely mimic human visual cognition, ultimately leading to more advanced and capable models. Additionally, we observe that current open-source models achieve matrix reasoning performance very close to that of closed-source models. However, VLMs face challenges in supporting multiple images as input and managing visual memory. Addressing these challenges is a crucial direction for building more robust open-source VLMs in the future.

\section{Ethics Discussion}
\label{appendix:ethics_discussion}

This research aims to advance LLMs and VLMs by providing a new benchmark for evaluating AI capabilities in visual reasoning. MaRs-VQA builds on the MaRs-IB (Attribution-NonCommercial 3.0 License), RAVEN (GPL-3.0 License). All code and data are available on GitHub. No conflicts of interest exist among the study’s contributors. The annotation process is IRB approved by a clinical institute.

\subsection{Negative Societal Impacts}
We foresee no direct negative societal impacts from our matrix reasoning benchmark. However, it could be misunderstood or misinterpreted as comparing AI ``thought'' to human cognition or misused to evaluate human abilities across demographics or ethnicity. We strongly caution against such misuse, as our datasets are not validated for human assessment.

Another concern relates to the future conclusion from our benchmark. While matrix reasoning is a crucial test for evaluating human intelligence, observing that VLMs with large model weights perform better on matrix reasoning tasks does not imply that the intelligence of MLLMs follows the same ``scaling law'' from the general domain. A comprehensive intelligence test requires accurate assessment using human-based tools, of which matrix reasoning is only one critical component. We cannot conclude that larger MLLMs can achieve human intelligence.

Additionally, there is a potential concern for discrimination against certain groups based on race, gender, or age in human study results. Although all human results in our experiment tables are sourced from previously published papers, we cannot guarantee that all previous research adhered to strict standards ensuring the inclusion of all groups in the human investigation process.

\subsection{Mitigating Bias and Negative Societal Impacts}

While the use of MaRs-VQA come with potential negative social impacts, there are viable mitigations that can address these concerns. These include adding instructions for proper use and restricting unethical human investigations. Users must be aware of the ethical implications associated with our benchmark and take appropriate measures to ensure its safe and responsible utilization.

\end{document}